\documentclass[journal,onecolumn,draftnocls,12pt,twoside]{IEEEtran}
\usepackage{amsmath,amssymb,amsfonts}
\usepackage{algorithm,algorithmic}
\usepackage{graphicx}
\usepackage{textcomp}
\usepackage{epsfig}
\usepackage{graphicx}
\usepackage{enumitem} 
\usepackage{subcaption}
\usepackage{hyperref}

\ifCLASSOPTIONcompsoc
  \usepackage[nocompress]{cite}
\else
  \usepackage{cite}
\fi

\begin{document}

\title{{TW-SMNet: Deep Multitask Learning of Tele-Wide Stereo Matching}}

\author{{Mostafa El-Khamy, Haoyu Ren, Xianzhi Du, and Jungwon Lee} \\
SOC Multimedia R\&D, Samsung Semiconductor Inc. (SSI), San Diego, CA 92121}

\markboth{}%
{{\textit{El-Khamy \MakeLowercase{et al.}}}, TW-SMNet: Deep Multitask Learning of Tele-Wide Stereo Matching}

\IEEEtitleabstractindextext{%
\begin{abstract}
In this paper, we introduce the problem of estimating the real world depth of elements in a scene captured by two cameras with different field of views, where the first field of view
(FOV) is a Wide FOV (WFOV) captured by a wide angle lens, and the second FOV is contained in the first FOV and is captured by a tele zoom lens.
 We refer to the problem of estimating the  inverse depth for the union of FOVs, while leveraging the stereo information in the overlapping FOV, as Tele-Wide Stereo Matching (TW-SM). We propose different deep learning solutions to the TW-SM problem. Since the disparity is proportional to the inverse depth, we train stereo matching disparity estimation (SMDE) networks to estimate the disparity for the union WFOV.  We further propose an end-to-end deep multitask tele-wide stereo matching neural network (MT-TW-SMNet), which simultaneously learns the SMDE  task for the overlapped Tele FOV and the single image inverse depth estimation (SIDE) task for the WFOV. Moreover, we design multiple methods for the fusion of the SMDE and SIDE networks. We evaluate the performance of TW-SM on the popular KITTI and SceneFlow stereo datasets, and demonstrate its practicality by  synthesizing  the Bokeh effect on the WFOV from a tele-wide stereo image pair.  
\end{abstract}

}

\maketitle

\IEEEdisplaynontitleabstractindextext

\IEEEpeerreviewmaketitle

\begin{figure}
\centering
\includegraphics[width=0.95\linewidth]{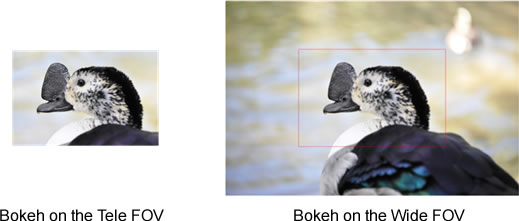}
\caption{Demonstration of the Bokeh effect on the central Tele FOV only (left) and on the full Wide FOV (right). \label{fig:Bokeh}}
\end{figure}

{\section{Introduction}\label{sec:introduction}}
\IEEEPARstart{D}{epth} estimation of real world elements in a scene has many applications in computer vision, scene understanding, image and video enhancement,  autonomous driving, simultaneous localization and mapping, and 3D object reconstruction. For example, accurate depth estimation allows accurate foreground-background segmentation, and hence the separation of the foreground (close) objects of interest from the background (far) objects in a scene. Foreground-background segmentation can be used in object detection, tracking, and image Bokeh. Bokeh is the soft out-of-focus blur of the background which can be mastered by using the right settings on expensive cameras with fast lens and wide apertures. Moreover, achieving the Bokeh effect by capturing with a shallow depth-of-field often requires the camera to be physically closer to the subject of interest, and the subject to be further away from the background. However, accurate depth estimation also allows one to synthesize the images captured by non-professional photographers or cameras with smaller lenses (such as mobile phone cameras) to obtain more aesthetically pleasant effects such as Bokeh (see Fig. \ref{fig:Bokeh}). Other applications of accurate depth estimation include 3D object reconstruction and virtual reality applications, where it may be desired to alter the background regions or the subject of interest to render them according to the desired virtual reality. For autonomous driving applications, where safety is most important, the depth estimated from the images captured by the cameras can be fused with those obtained from other sensors to improve the accuracy of estimating the distance of the detected objects from the camera.

\begin{figure*}[t!]
    \centering
\begin{subfigure}[b]{0.75\textwidth}
\includegraphics[width=0.48\textwidth]{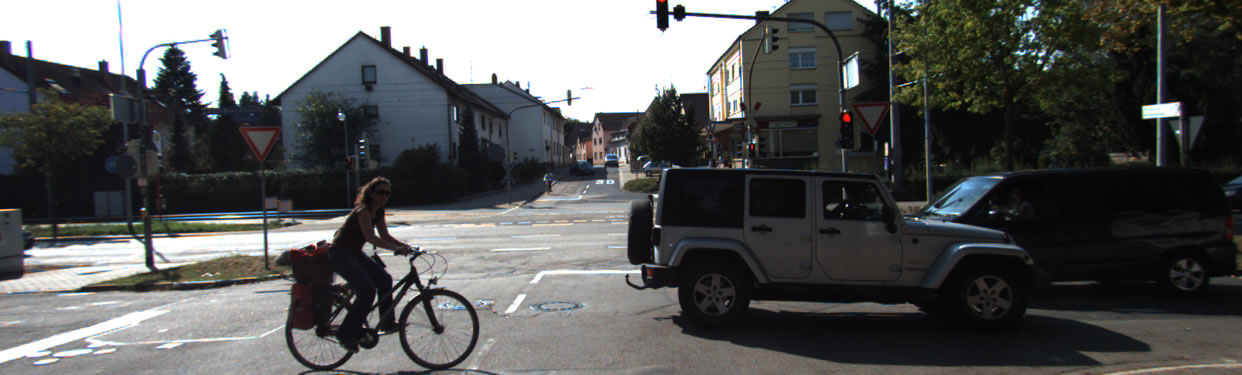}
\hfill
\includegraphics[width=0.48\textwidth]{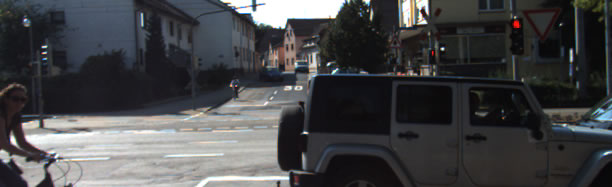}
\caption{Example of a tele-wide stereo input with a left Wide FOV and a right Tele FOV. }
\label{fig:tele_wide_kitti}
 \end{subfigure}		
    \\
\begin{subfigure}[b]{0.75\textwidth}
\includegraphics[width=\textwidth]{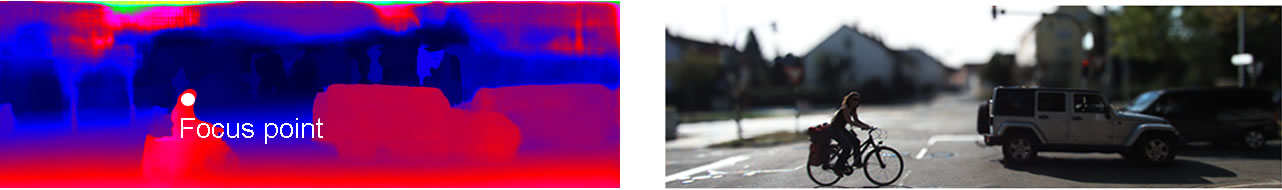}
\caption{The estimated tele-wide disparity map and the synthesized full-FOV Bokeh.  }
\label{fig:tele_wide_kitti2}
 \end{subfigure}
\caption{Tele-wide stereo matching on a KITTI example, to generate Wide FOV Bokeh)}
\label{fig:kitti2015intro_Bokeh}
\end{figure*}

The problem of estimating depth in an image from two stereo cameras with an identical field of view (FOV) has been well studied~\cite{scharstein2002taxonomy, SGM,costvolumefiltering}.  Depth estimation from two stereo rectified images can be obtained by calculating the disparity (which is the horizontal displacement) between matching pixels in both images. This is often done by stereo matching techniques that find the corresponding points in both images. Stereo matching involves the extraction of features from the stereo images and the computation of a cost volume to match between the features along the same horizontal line in the left and right rectified images using different similarity measures. This is then followed by aggregation of the cost metrics to optimize the disparity estimation.

Multi-camera systems, where more than two cameras with the same focal lengths are deployed at different baseline lengths and directions, have been proposed \cite{kanade1995development}. In this multi-baseline system, the cameras are assumed to have the same focal length. Hence, multiple disparity measurements for the common (overlapping) field of view are estimated by stereo pairs with multiple baselines, and then fused to obtain a statistically more accurate depth map. Recently, there has been a renewed interest in multi-camera system deployment in the autonomous driving systems, as well as in mobile handsets. To make the best benefit of the extra cameras, the camera lenses are chosen to have different focal lengths in order for the device to have good resolution at both near and far objects, which results in different field of views (FOVs). For example, recent cell phones are equipped with two or more cameras.  Example specifications of the cameras are ($f/1.7$, $26 mm$, $1 \times$ optical zoom) and ($f/2.4$, $52 mm$, $2 \times$ optical zoom), respectively. One reason for having cameras with different focal lengths is for diversity,
where the first camera has a wider aperture for better light sensitivity, and the second camera has a longer focal length for capturing a higher image resolution at twice the optical zoom. 

We consider in this work a two camera system, where the first camera has $1 \times$ the optical zoom, and the second camera has $2 \times$ the optical zoom. 
We call the field of view of the first camera with $1 \times$ the zoom, as the Wide FOV (WFOV). The field of view of the second camera with twice the optical zoom is the central part of the WFOV, and we call it the Tele FOV (TFOV). 

We refer to the problem of the estimation of the inverse depth for the region described by the union of two different FOVs, while leveraging the stereo information from the overlapping FOV, as tele-wide stereo matching (TW-SM).  
To the best of our knowledge, this work is the first to consider the TW-SM problem.
Since the disparity is directly proportional to the inverse depth, we also refer to this problem as disparity estimation of the union of stereo FOVs. 
 One application of stereo  matching is to produce a Bokeh effect in the image, by blurring the background, while keeping the object of interest in focus. With TW-SM, the Bokeh effect can be synthesized for the full WFOV image, rather than for the central TFOV region that is achievable by conventional stereo matching. 
Fig.~\ref{fig:Bokeh} demonstrates the big difference between the Bokeh effect on the full WFOV and on the central TFOV only. Fig.~\ref{fig:kitti2015intro_Bokeh} demonstrates the result achievable by this paper on an example from the KITTI stereo dataset, where the disparity map is estimated using TW-SM and used to synthesize the Bokeh effect on the WFOV.

\begin{figure*}[t!]
\centering
\includegraphics[width=0.75\linewidth]{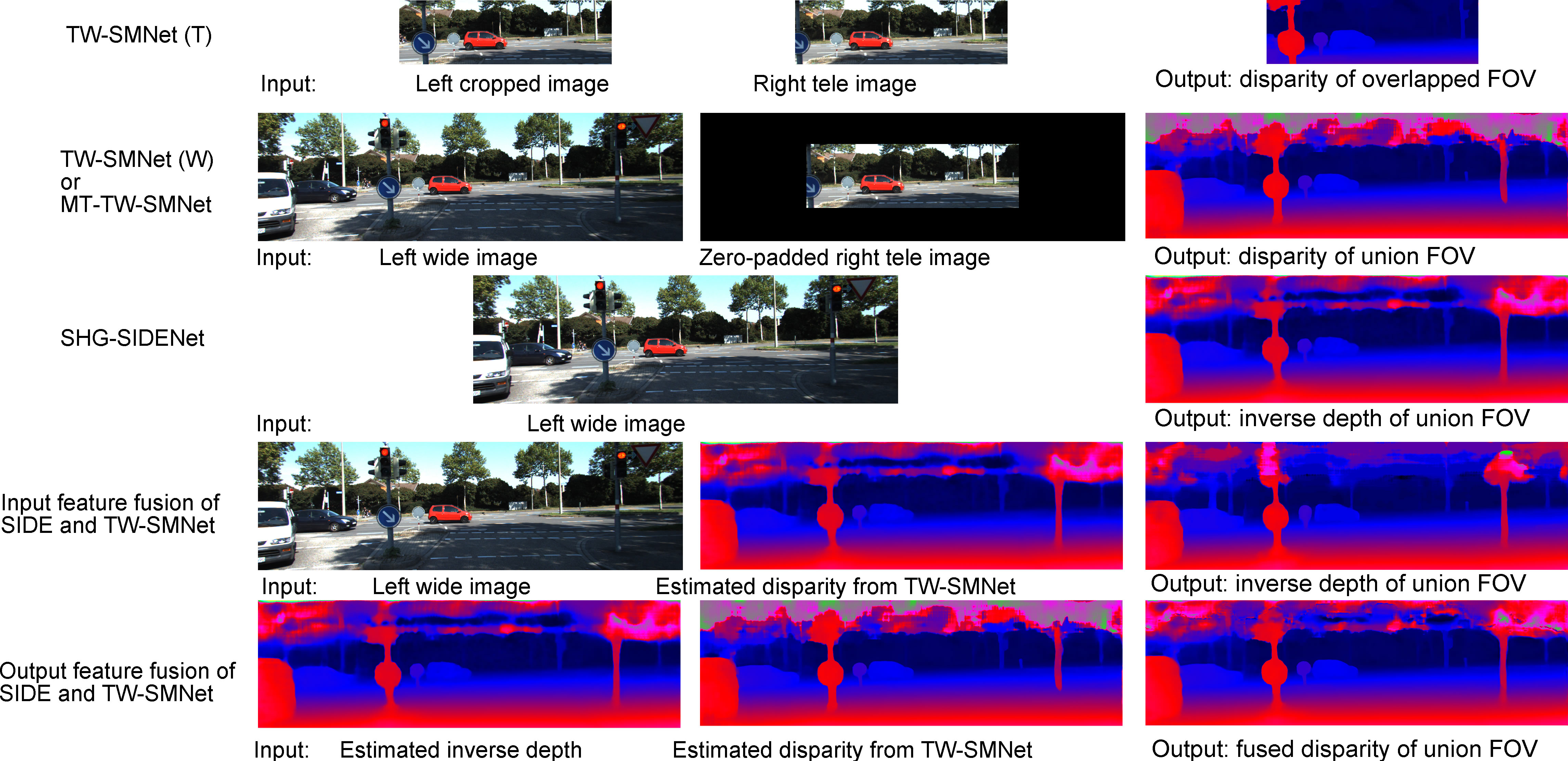}
\caption{The proposed solutions for tele-wide stereo matching,  and their input and output features.}
\label{fig:examplefournets}
\end{figure*}

  We consider multiple deep-learning based approaches to the tele-wide stereo matching problem. First, we develop tele-wide stereo matching networks (TW-SMNets) that attempt to do stereo matching between the tele and wide images to estimate the disparity. Hence,  we train TW-SMNet(T) to estimate a disparity map  for the overlapped TFOV, and train TW-SMNet(W) to estimate a disparity map for the union WFOV. Second, we pose the problem as a single-image inverse-depth estimation (SIDE) problem by formulating the disparity as the scaled inverse depth. Hence, we train deep SIDE neural networks to estimate the inverse depth from the wide FOV image only. Third, we design a deep multitask tele-wide stereo matching neural network (MT-TW-SMNet) which takes the stereo input images with different FOVs, and estimates the disparity for the overlapped FOV using the stereo input, as well as the inverse depth for the union FOV using the WFOV input image \footnote{Our preliminary multitask tele-wide stereo matching results have been accepted for publication at the IEEE Conference on Image Processing, ICIP 2019 \cite{elkhamy_MTTW}.}. Fourth, we design multiple methods for the fusion of these two approaches, SIDE and TW-SM. For example, we consider input feature fusion such as RGBD SIDE, where the SIDE network uses the disparity estimates from the multitask stereo matching network as additional input features, to guide its estimation of inverse depth.  We also consider output feature (decision) fusion, where we design a fusion network to fuse the disparity estimates from the tele-wide stereo matching network and the WFOV single image inverse-depth estimation network.  
Examples of the inputs and outputs of the proposed networks are shown in Fig. \ref{fig:examplefournets}.

The rest of this paper is organized as follows. Section~\ref{sec:SM} explains the preliminaries and the related works to stereo disparity estimation and single image depth estimation. In Section~\ref{sec:TWDE}, our different proposals for tele-wide stereo matching are explained. Section~\ref{sec:perf} gives the experimental results of tele-wide networks on KITTI and SceneFlow datasets. Section~\ref{sec:twfusion} discusses our proposed input feature fusion and output feature fusion between the different tele-wide disparity estimation methods. Section~\ref{sec:conc} concludes the paper.

\section{Preliminaries \label{sec:SM}}
\subsection{Stereo disparity estimation} 

Stereo disparity estimation techniques involve feature extraction, matching cost computation, disparity aggregation and computation, and  disparity refinement \cite{scharstein2002taxonomy}.  Matching cost computation at a given disparity is based on measuring the similarity between pixels in the left and right images at this disparity shift. The cost computation can simply be the sum of absolute differences of pixel intensities at the given disparity \cite{PAMI_SMC}.  Disparity aggregation can be done by simple aggregation of the matching cost over local box windows, or by guided-image cost volume filtering \cite{costvolumefiltering}. Estimation of the disparity calculation can be done by local, global, or semiglobal methods. Semiglobal matching (SGM) \cite{SGM} methods are less complex than global methods such as graph cut algorithms \cite{kolmogorov2004energy}. Semiglobal matching is also more robust than local window-based methods. SGM performs cost aggregation by approximate minimization of a two dimensional energy function towards each pixel along eight one dimensional paths. Disparity refinement is classically done by further checking for left and right consistencies, invalidating occlusions and mismatches, and filling such invalid segments by propagating neighboring disparity values.

Deep learning approaches to solve the disparity estimation problem surged after the significant efforts in collecting datasets with stereo input images and their ground truth disparity maps, e.g. SceneFlow~\cite{sceneflow}, KITTI 2012~\cite{kitti2012}, KITTI 2015~\cite{kitti2015}, and the Middlebury~\cite{middlebury} stereo benchmark datasets. The existence of such datasets enabled supervised training of deep neural networks for the task of stereo matching,  as well as the transparent testing and benchmarking of different algorithms on their hosting servers.

 Convolutional neural networks (CNNs) are now widely investigated to solve problems in computer vision  and image processing.  Similar to the classical disparity estimation techniques, CNN-based disparity estimation involves feature extraction, matching cost estimation, disparity aggregation and computation, and disparity refinement. First, deep features are extracted from the rectified left and right images using deep convolutional networks such as ResNet-50~\cite{resnet} or VGG-16~\cite{vgg}. The cost volume is formed by measuring the matching cost between the extracted left and  right deep feature maps at different disparity shifts. Typical choices for the matching cost are simple feature concatenation, or calculation of metrics such as absolute distance or correlation~\cite{sceneflow,MCCNN,flownet,luo} or by extending the cost volume to compute multiple metrics~\cite{amnet}. The cost volume is further processed and refined by a disparity computation module that regresses to the estimated disparity. Refinement networks can then be used to further refine the initial coarse depth or disparity estimates.

 Zbontar et. al \cite{zbontar2016stereo} designed a deep Siamese network for stereo matching, where the network was trained to predict the similarity between image patches. Luo et al. \cite{luo2016efficient} improved this work \cite{zbontar2016stereo} and  made it more efficient, where they formulated the matching cost computation  as a multi-label classification problem. Shaked et al. \cite{shaked2017improved} proposed a highway network for the matching cost computation and a global disparity network for the prediction of disparity confidence scores. 
Other studies focused on the post-processing of the disparity map for disparity refinement.  Seki et al. further extended SGM to SGM-Net which deployed SGM penalties, instead of manually-tuned penalties, for regularization. Post-processing networks were proposed  to detect incorrect disparity and  replace them by others estimated from the local regions, and further refined by a post-refining module \cite{gidaris2017detect}. 
More complicated convolutional neural networks are utilized in later approaches, including SharpMask \cite{pinheiro2016learning}, RefineNet \cite{liang2017learning},  the label refinement network \cite{islam2017label}, the stacked hourglass (SHG) architecture  \cite{fu2017stacked}\cite{newell2016stacked}, and the atrous multiscale architecture which does dense multiscale contextual aggregation~\cite{amnet}. 

Different neural network architectures were proposed for stereo matching.
There are two main approaches, the encoder-decoder architectures and the spatial pyramid  pooling architecture. The encoder-decoder utilizes a cascade architecture, which integrates top-down and bottom-up information through skip connections. The fully convolutional network (FCN) \cite{long2015fully} was one of the pioneering works following this architecture to aggregate coarse-to-fine predictions to improve the segmentation quality.
Mayer et al. \cite{mayer2016large} designed an end-to-end network (DispNet) for both the disparity estimation and optical flow estimation. Pang et al. \cite{pang2017cascade} developed a two-stage network based on DispNet, where the disparity map is calculated at the first stage, its multiscale residual map is extracted at the second stage, and the outputs of these two stages are further combined. Kendall et al. \cite{kendall2017end} introduced GC-Net, where 3D convolutional layers are designed for cost volume regularization using 3D convolutions. 
These end-to-end networks exploit multiscale features and hierarchical relationships between the earlier and later neural network layers for disparity estimation. The contextual information is utilized to reduce the mismatch at ambiguous regions, and improve the depth estimation. 
 Spatial pyramid pooling (SPP) integrates  deep receptive fields with multiscale context information. Global pooling with FCN \cite{liu2015parsenet} has been shown to enlarge the empirical receptive field to extract information from the whole image. Later works improved the SPP by introducing the atrous (dilated) convolutions. Zhao et al. designed PSPNet \cite{zhao2017pyramid}, where a pyramid pooling module is adopted to generate a multiscale contextual prior. Chen  et al. proposed the atrous spatial pyramid pooling (ASPP) module in DeepLab v2 \cite{chen2018deeplab} for multiscale feature embedding.  DeepLab v3 \cite{chen2017rethinking} introduced a newly designed ASPP module, where global feature pooling and parallel dilated convolutions with different dilation factors are utilized to aggregate features from different receptive fields. 

\begin{figure}[t!]
\centering
\includegraphics[width=0.9\linewidth]{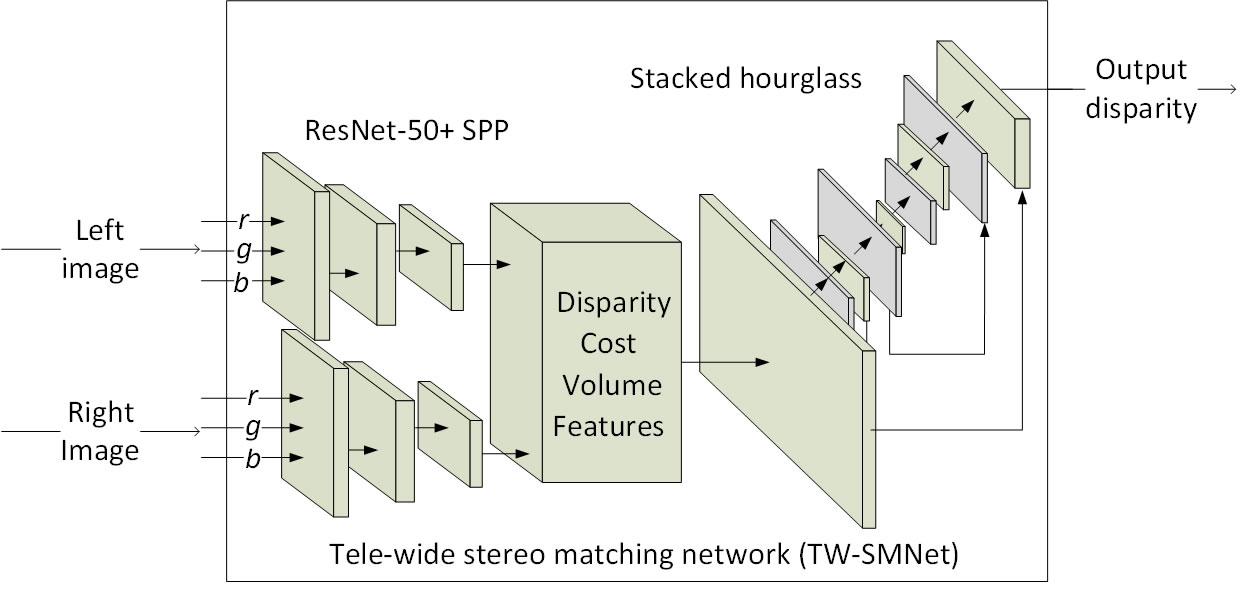}
\caption{Architecture of the tele-wide stereo matching network (TW-SMNet). \label{fig:TW-SMNet}}
\end{figure}

Chang et al. proposed PSMNet \cite{chang2018pyramid}, a pyramid stereo matching network consisting of two main modules: spatial pyramid pooling and 3D CNN. In PSMNet, the spatial pyramid pooling module takes advantage of the capacity of global context information by aggregating contexts at different scales and locations to form a cost volume. The 3D CNN learns to regularize the cost volume using stacked multiple hourglass networks in conjunction with intermediate supervision. In this paper, our proposed tele-wide stereo matching network uses PSMNet as a baseline of the stereo matching network TW-SMNet, and its architecture is demonstrated in Fig.~\ref{fig:TW-SMNet}.

\subsection{Single image depth estimation}
Previous approaches for depth estimation from single images can be categorized into three main groups: (i) methods operating on hand crafted features, (ii) methods based on graphical models and (iii) methods adopting deep neural networks. Earlier works addressing the depth prediction task belong to the first category. Hoiem et al. \cite{hoiem2005automatic} introduced photo pop-up, a fully automatic method for creating a basic 3D model from a single photograph. Karsch et al. \cite{karsch2014depth} developed Depth Transfer, a non parametric approach where the depth of an input image is reconstructed by transferring the depth of multiple similar images and then applying some warping and optimizing procedures. Ladicky \cite{ladicky2014pulling} demonstrated the benefit of combining semantic object labels with depth features. Other works exploited the flexibility of graphical models to reconstruct depth information. One of the most commonly-used technologies is the Conditional Random Field (CRF). By defining different unary and pairwise potentials, we can have different CRFs. For instance, Delage et al. \cite{delage2006dynamic} proposed a dynamic Bayesian framework for recovering 3D information from indoor scenes. A discriminatively-trained multiscale Markov Random Field (MRF) was introduced in \cite{saxena20083}.

More recent approaches for depth estimation are based on CNNs. The pioneering work by Eigen \cite{eigen2014depth} introduced a two-scale architecture consisting of the coarse-scale network and the fine-scale network. The coarse-scale network is a convolutional neural network that identifies the global scene context. This coarse depth map, along with the original input image are then fed to the fine-scale fully convolutional network to refine the depth result. The scale-invariant loss is utilized in this work, which further improves the robustness of the estimated depth map. This work is further extended in \cite{eigen2015predicting}, where the depth estimation, surface normal estimation, and semantic segmentation are integrated into one unified network. Li et al. \cite{li2017two} considered using a loss function with components in both the depth domain and the gradient of depth domain, yielding a two-channel network, to learn both the depth map and its gradient which are fused together to generate the final output.

There are a few works who considered combining the CNN and CRF together.  Conventional CRF-based depth estimation used the RGB image as the observation.  Mousavian et al.~\cite{mousavian2016joint} proposed to combine depth estimation and semantic segmentation together. First, a CNN is utilized to extract feature maps of the depth and semantic labels at the same time. Then, these feature maps are fused as the input of a CRF. Xu et al.~\cite{xu2017multi} proposed a cascade structure for CNN-CRF depth estimation, where a side output of the CNN is used  as the input to the CRF for depth estimation at a certain scale, and then estimates are refined at subsequent levels. Two CRF architectures, multiscale CRF (one CRF uses all the feature maps in the potential function), and cascade CRF (the depth estimated by  one CRF is the input to the next CRF) are proposed, where the CRFs are implemented as neural network modules to enable end-to-end training.

One disadvantage of the above works is that they are not fully convolutional since a fully connected layer is still utilized for depth regression. This significantly increases the model complexity and the computational burden, which prompted Laina et al.~\cite{laina2016deeper} to propose a fully convolutional network (FCN). FCN is based on ResNet-50 with additional up-sampling layers to recover the loss in resolution from the pooling layers, and is trained with the BerHu loss. 
Laina's work was extended so the input constitutes of a randomly sampled sparse depth map, in addition to the input RGB features of an image \cite{ma2017sparse}. As expected, the additional prior information about the sparse depth samples improves the SIDE accuracy significantly. 
 Li et al.~\cite{li2017single} reformulated the SIDE problem as a classification problem by quantizing the depth in the log space and using a soft-weighted-sum inference. DORN~\cite{fu2018deep} further formulated the depth estimation problem as an ordinal regression problem and achieved  state-of-art performance using atrous spatial pyramid pooling (ASPP) modules to capture multiscale information.

\section{Tele-wide Stereo Matching \label{sec:TWDE}}
Given two stereo cameras, the stereo matching techniques described in Section \ref{sec:SM} can be utilized to estimate the disparity map for the overlapping FOVs. However, to our knowledge disparity estimation for the union of FOVs has not been well studied or benchmarked previously in the literature.
Here, we consider the special case when the FOV of one camera is contained inside that of the other camera. The practical need for this case has surfaced recently due to the equipment of recent mobile devices with stereo cameras (tele camera and wide camera) which have different focal lengths, and where the tele camera is specified to capture the center FOV of the other camera at twice the optical zoom.

In this section, we propose different deep learning solutions to solve the tele-wide disparity estimation problem. We deploy a tele-wide stereo-matching network (TW-SMNet) to work on the left wide-image and the right tele-image to generate a disparity map for the wide FOV.  This network is expected to perform well on the TFOV. We also benchmark the performance of this tele-wide network when trained to learn the disparity map for the full WFOV. Next, we  modify the TW-SMNet to be a single-image inverse depth estimation network (SIDENet), which works only on the left wide-image to generate the inverse depth prediction for the full WFOV. Based on observations from these networks, we propose a multitask tele-wide stereo matching network (MT-TW-SMNet) which concurrently learns to do stereo matching and single image inverse depth estimation, to get better disparity estimates for the full WFOV.

\begin{figure}[t!]
\centering
\includegraphics[width=0.9\linewidth]{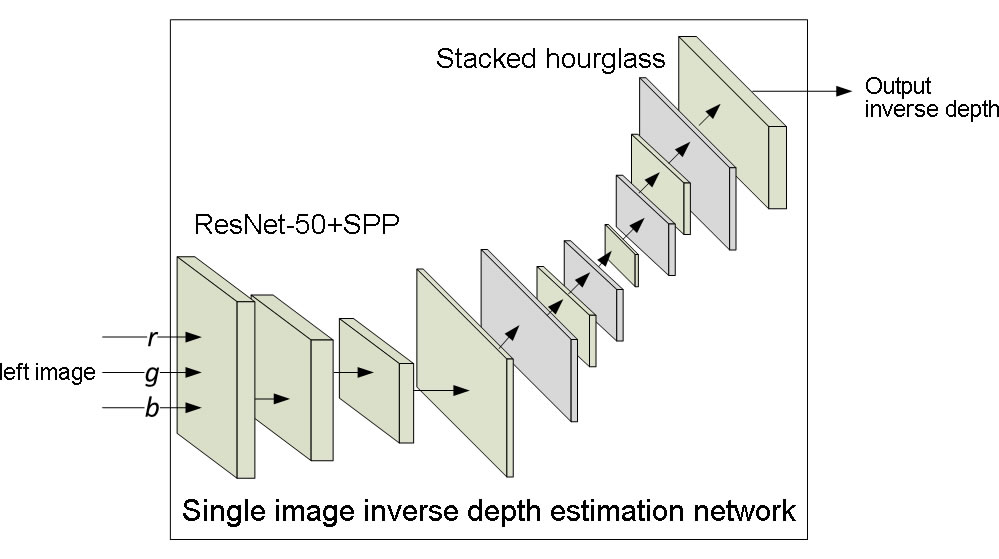}
\caption{Architecture of the stacked hourglass single image depth estimation network (SHG-SIDENet). \label{fig:SIDENet}}
\end{figure}

\subsection{Tele-Wide Disparity Estimation \label{sec:tw}}
We establish a baseline solution to the tele-wide disparity estimation problem by casting the problem as a standard stereo matching problem. We implement a tele-wide stereo matching network (TW-SMNet) that takes as input a tele-wide stereo pair, representing the left WFOV and the right TFOV, cf. Fig.~\ref{fig:tele_wide_kitti}. We train two different TW-SMNets; TW-SMNet(T) which is trained to output an estimated disparity map for the overlapping TFOV only, and TW-SMNet(W) which is trained to estimate the disparity for the full WFOV. 
TW-SMNet(T) and TW-SMNet(W) modify the tele-wide stereo inputs differently according to the different training objectives, as demonstrated in Fig.~\ref{fig:examplefournets}. The input features of TW-SMNet(W) uses a wide stereo image pair, where the right tele image is zero-padded after down-sampling to match the resolution and size of the left wide image. TW-SMNet(T) uses a tele stereo image pair, where its input features are the right tele image and the left wide image after it is center cropped for the TFOV and up-sampled to match the size and resolution of the right tele image.

The network architecture of the TW-SMNets is shown in Fig.~\ref{fig:TW-SMNet}. The TW-SMNet uses a similar architecture to PSMNet~\cite{chang2018pyramid}, and consists of four main parts: feature extractor, cost volume for left/right feature map matching, stacked hourglass module, and disparity regression. Feature extraction is done by 3 $3\times 3$ convolutional layers followed by a $50$-layer residual network (ResNet-50) constituting of multiplicities of  four different residual blocks with atrous convolutional layers. This is followed by a spatial pyramid pooling (SPP) module that extracts hierarchical context information at multiple scales. The SPP module is composed of four average pooling blocks of fixed-sizes $[64\times 64, 32\times 32, 16\times 16, 8\times 8]$. After feature extraction, a cost volume is utilized to learn the matching costs between the left and right feature maps, where the left feature maps are aligned and concatenated with their corresponding right feature maps at each disparity shift. This results in a four dimensional feature map of dimension (number of channels$\times$maximum disparity$\times$height$\times$width). Hence a 3D CNN is used for cost aggregation, where three hourglass modules are stacked  to learn  high level local contextual information while keeping the low level global contextual information. 

The TW-SMNet is trained with a classification-based regression loss, where the network regresses the feature maps at the output of the stacked hourglass to a continuous disparity map.  
The disparity is classified into $D$ bins whose values are the integers from 1 to $D$, where $D$ is the maximum disparity. The classification probabilities at the $D$ bins are calculated for the disparity at each pixel. The expected disparity is then calculated from the estimated  disparity probabilities. The predicted disparity for each pixel location is given by \eqref{eq:regression}, 
\begin{equation}
d_i=\sum_{j=1}^D{j\times p_i^j},
\label{eq:regression}
\end{equation} 
where $d_i$ is the estimated disparity of pixel $i$,  $p_i^j$ represents the soft probability of the disparity $d_i$ falling in the bin with value $j$. 
For robust regression, the Huber loss ($H_{1}$) is used to measure the difference between the predicted disparity $d_i$ and the ground truth disparity $d_i^{gt}$, as shown in ~\eqref{eq:loss},
\begin{equation}
L(d_i,d_i^{gt})=\frac{1}{N}\sum_i {H_1} (d_i-d_i^{gt}),
\label{eq:loss}
\end{equation} 
where $N$ is the total number of labeled pixels, and the Huber loss at any $\delta$ is defined by~\cite{huber1973robust}
$$ {H_\delta}(x)= \begin{cases} 
      0.5 x^2, & \mbox{if}~|x| \leq \delta, \\
      \delta|x|-0.5 \delta^2, & \mbox{otherwise}.  \end{cases} $$

On the one hand,  TW-SMNet(T) only provides disparity estimates for the overlapping TFOV based on stereo matching at the tele regions only, and it should be the most accurate within the TFOV. 
On the other hand, TW-SMNet(W) estimates the disparity for the full WFOV including the region surrounding the TFOV. Since TW-SMNet(W) has the same input information as TW-SMNet(T) in the tele region, the accuracy of TW-SMNet(W) should be close to that of TW-SMNet(T) in the TFOV. However, due to the lack of stereo matching information  at the surrounding region in the wide stereo input, TW-SMNet(W) suffers a relatively larger error when estimating the disparity for the pixels in the surrounding region, which are the pixels in the Wide FOV but not in the TFOV.

\subsection{Wide FOV Inverse Depth Estimation}
\begin{figure}[h!]
\centering
\includegraphics[width=0.5\linewidth]{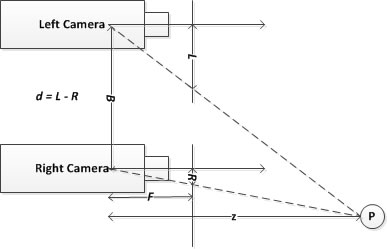}
\caption{Calculation of depth $z$ from stereo disparity $d$, focal length $F$, and camera baseline $B$. \label{fig:stereo}}
\end{figure}

To improve the performance on surrounding region, we propose a stacked hourglass single image inverse depth estimation network (SHG-SIDE), which estimates an inverse depth map from the left wide RGB image only.
 For each pixel, the computed disparity $d$ is proportional to the `inverse depth' $\zeta$, where the depth $z$ (distance to the scene point) is related to the disparity by the camera baseline  $B$ (distance between the two camera centers), and the camera focal length $F$, as demonstrated by Fig. \ref{fig:stereo}, and  
\begin{equation} 
\zeta = \frac{1}{z}=\frac{d}{FB}. 
\label{eq:inv_depth}
\end{equation}

Since, an inverse depth map can be converted to a disparity map by knowledge of the camera baseline and focal length which are predetermined constants for a specific camera setup, the SHG-SIDE network is trained to regress to the disparity (i.e. a scaled inverse depth) directly. The last convolutional layer of SHG-SIDENet is set to have $D$ output neurons for classification, and the same classification-based robust regression as TW-SMNet is used for training of the SHG-SIDE network and the prediction of the inverse-depth maps.   
The network architecture of SHG-SIDENet is modified from that of TW-SMNet. We remove the feature extraction channel for the right tele image and the cost volume. All the three dimensional convolutional layers after the cost volume are hence shrunk to a two dimensional form. The network structure is shown in Fig.~\ref{fig:SIDENet}. 
Since the SHG-SIDENet estimates the inverse depth by  scene understanding it can provide more accurate depth estimates than TW-SMNet in the surrounding region where the stereo matching information is incomplete.

\begin{figure}[t!]
\centering
\includegraphics[width=\linewidth]{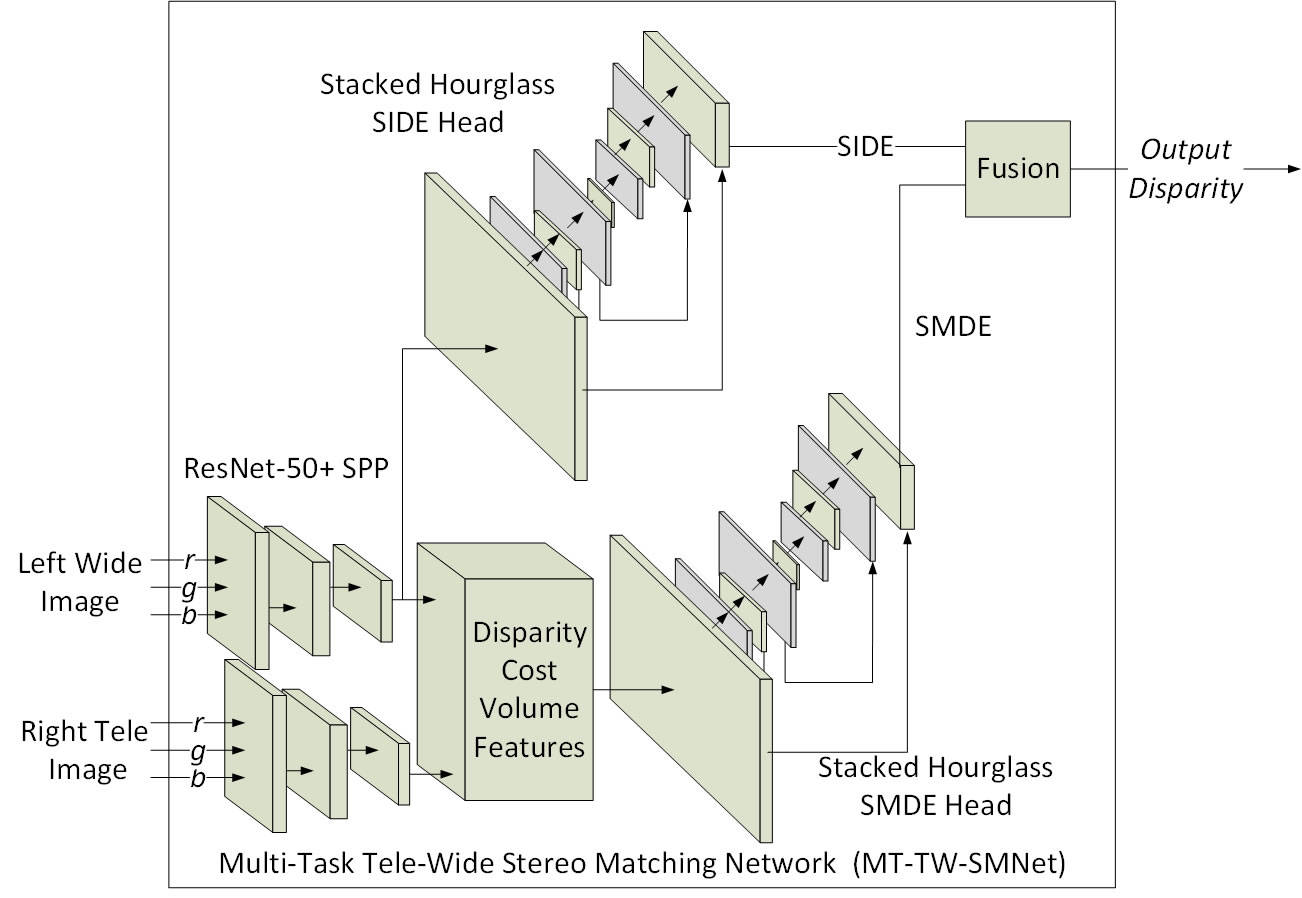}
\caption{The network architecture of the multitask tele-wide stereo matching network (MT-TW-SMNet). \label{fig:MT-TWSMNet}}
\end{figure}

\begin{table*}
\caption{Error rate of tele-wide disparity estimation networks on the KITTI stereo 2015 validation dataset; `cen' stands for the `center' TFOV region, and `sur' stands for the region in the WFOV `surrounding' the TFOV. }
\small
\renewcommand{\arraystretch}{1.0}
\centering
\begin{tabular}{|c|c|c|c|c|c|c|}
\hline
Model name & type & left image & right image & error-all(\%) & error-cen(\%) & error-sur (\%)  \\ \hline
DORN & single image & wide & N/A & 17.55 & 11.96 & 20.68 \\  
SHG-SIDENet &  single image & wide & N/A & 12.62 & 7.31 & 15.80 \\
TW-SMNet(T) &  stereo  & tele & tele & N/A & 1.68 & N/A\\
TW-SMNet(W) &  stereo  & wide & tele & 13.10 & 1.86 & 19.63 \\
MT-TW-SMNet& stereo & wide & tele & 12.70 & 1.94 & 18.99  \\ \hline
\end{tabular}
\label{table:1nKITTI}
\end{table*}

\begin{table*}
\caption{End-point error of tele-wide disparity estimation networks on the SceneFlow test set.}
\small
\renewcommand{\arraystretch}{1.0}
\centering
\begin{tabular}{|c|c|c|c|c|c|c|}
\hline
Model name & type & left image & right image & error-all(pixels) & error-cen(pixels) & error-sur (pixels)  \\ \hline
SHG-SIDENet &  single image & wide & N/A &7.47 & 8.32 & 7.20 \\
TW-SMNet(T) &  stereo  & tele & tele & N/A & 1.28 & N/A\\
TW-SMNet(W) &  stereo  & wide & tele &5.79 & 1.88 & 7.10 \\
MT-TW-SMNet & stereo & wide & tele & 5.61 & 1.62 & 7.08  \\ \hline
\end{tabular}
\label{table:1NSceneFlow}
\end{table*}

\subsection{Multitask Network for Tele-Wide Stereo-Matching}
We investigated both single image inverse depth estimation (SIDE) and stereo matching (SM) solutions to the tele-wide disparity estimation problem. 
Whereas the TW-SMNet has an advantage in the tele FOV, the SHG-SIDENet has an advantage in the surrounding region of the wide FOV. This inspires us to propose an end-to-end multitask tele-wide stereo matching network (MT-TW-SMNet) which combines the TW-SMNet and the SIDENet together. The proposed MT-TW-SMNet takes as input both the left wide image and the zero-padded right tele image. The ResNet-50 followed by SPP, as explained in Sec.~\ref{sec:tw}, are used as a feature extractor which is shared between both the SM and SIDE tasks. As shown in  Fig.~\ref{fig:MT-TWSMNet},  the MT-TW-SMNet is a two-branch network, where the SM branch constructs a cost volume from the features extracted from the left and wide images for disparity computation, and the SIDE branch uses the  features extracted from the left wide image for scene understanding. The loss function of MT-TW-SMNet is a linear combination of the classification-based robust regression losses used to train the SIDENet and TW-SMNet, as computed by~\eqref{eq:multask}
\begin{equation}
L(d_i,d_i^{gt})=L_{SMDE}(d_i,d_i^{gt})+\alpha L_{SIDE}(d_i,d_i^{gt}),
\label{eq:multask}
\end{equation}
where $\alpha$ is a constant that is empirically chosen as $1.0$ in this work.

In MT-TW-SMNet, the learning process of SIDE branch will assist the TW-SM branch to make a better understanding of the global context. Since both branches share the same feature extractor and are trained end-to-end, the accuracy of the stereo matching branch  will be improved. 
At inference time, the disparity estimates from both branches can be fused together. For example, the final disparity estimate can be a linear combination of that estimated by both the SIDE and SMDE
branches. However, for efficient inference, we only used the output of
the SMDE branch, where we found that the additional learning
of the SIDE objective enabled the SMDE branch to make
a better understanding of the global context and provide a
more accurate disparity estimate than TW-SMNet(W) in the
surrounding region, while preserving its stereo matching
accuracy in the tele region.

\section{Performance Evaluation of MT-TW-SMNet \label{sec:perf}}
\subsection{Datasets}
We use two popular stereo datasets to evaluate our tele-wide disparity estimation methods,  SceneFlow \cite{mayer2016large}, and the KITTI Stereo 2015 \cite{geiger2013vision}. The left image is used as the left-wide image, while the center-cropped region of the right image is used as the right tele image, which has half the width and half the height of the left wide image.
SceneFlow is a large scale synthetic dataset containing $35,454$ training and $4,370$ testing images of size $960\times540$. This dataset provides dense and elaborate ground truth disparity maps. The end-point disparity error is calculated among all pixels with valid disparity labels when evaluating the performance of our tele-wide networks.
KITTI is a real-world dataset with street views from a driving car. It contains training stereo image pairs with sparse ground-truth disparities obtained using LiDAR. The image size in KITTI is around $1242\times375$. We use the KITTI stereo $2015$ validation set ($40$ labeled image pairs), and test set ($200$ unlabeled image pairs) to evaluate the performance of our networks. Since only 200 training images are provided, we fine-tune our SceneFlow networks on these training images to obtain the KITTI models. In the evaluation,  a pixel to be correctly estimated if the disparity or flow end-point error is less than $3$ pixels or less than $<5\%$. We note that the disparity is not labeled for the sky regions in the KITTI training and testing datasets, and all non-labeled regions are ignored for evaluation purposes.

\subsection{Performance comparisons between TW-SMNet, SHG-SIDENet, and MT-TW-SMNet}
We evaluate the accuracy of the proposed tele-wide disparity estimation networks: the single image inverse depth estimation network SHG-SIDENet, the stereo matching network TW-SMNet, as well as the multitask network MT-TW-SMNet. In Table \ref{table:1nKITTI}, we give the error rates on the KITTI 2015 validation datasets, as well as the accuracy of state-of-art single image depth estimation network DORN \cite{fu2018deep}. Since the DORN provides an output depth map, we convert it to a disparity map when calculating the error. The center error is labeled `cen' and is the error when calculated over the pixels in the TFOV only, while the surrounding error is labeled `surr' and  is calculated over the pixels in the WFOV which are not in center TFOV region.  It can be seen that the tele-wide networks achieve better accuracy than DORN. We can find that the best center accuracy is achieved by TW-SMNet(T) which is based on stereo matching of the left-right tele image pairs. However, TW-SMNet(T) does not provide the surrounding disparity. TW-SMNet(W) uses the left wide image and the zero-padded right tele-image as input, and estimates a disparity map for the full WFOV, including the surrounding region. Moreover, the accuracy of TW-SMNet(W) is close to that of TW-SMNet(T) in the TFOV region.  In contrast, the single image inverse depth estimation network SHG-SIDENet gives a more accurate estimate for the surrounding disparity than TW-SMNet(W), but the central accuracy is worse. These results make sense because the stereo disparity estimation is a matching problem. If both the left RGB image and right RGB image are given, the accuracy will be good. In case of missing information, as in the surrounding region, the cost-volume computation of of TW-SMNet(W) will struggle to find the left-right correspondence, so  the estimation accuracy will not be as good as with single image inverse depth estimation which relies on scene understanding. 
We also observe that the center accuracy of the multitask network MT-TW-SMNet is a bit lower than that of TW-SMNet(W) and TW-SMNet(T) in the central TFOV region, but the surrounding accuracy and the overall-pixel accuracy are better. 

 Similar observations can be found by testing on the SceneFlow dataset, as shown in Table \ref{table:1NSceneFlow}. 
The end-point-error of our tele-wide disparity estimation networks on SceneFlow confirm that the MT-TW-SMNet acheives the the lowest error of $5.6$ pixels. It has better accuracy than SIDENet in the overlapping Tele FOV, and better accuracy than TW-SMNet(W), which is only trained for stereo matching, in the surrounding non-overlapping region.  This shows the effectiveness of our multitask learning strategy, where learning SIDE as an auxiliary task helped the main SM branch provide a better estimate in the surrounding region.

We also submitted our results to the KITTI evaluation server for testing on the  KITTI Stereo 2015 test set. KITTI evaluates the disparity error for all the pixels in the wide FOV, and also breaks down the error rates separately for the pixels belonging to the foreground objects  or to the background regions, as shown in Table \ref{table:2nKITTI}. The MT-TW-SMNet achieved better overall tele-wide disparity estimation accuracy than the SIDENet. Our multitask tele-wide disparity estimation also ranks on the KITTI leader board better than other methods which use full wide left and right images for stereo disparity estimation, such as \cite{hosni2013fast,richardt2016dense,pena2016disparity}.
The table also shows that with fusion between the different methods (MT-TW Fusion), the result can be further improved so the overall disparity error is only $11.96\%$\footnote{Detailed analyses and visualizations of the MT-TW-SMNet disparity maps on KITTI Stereo 2015 test set can be found at \cite{MTTWSMNet-kitti}.}. The different fusion methods are explained in the next section.

\begin{table}
\caption{Full FOV disparity error rate (\%) on KITTI Stereo 2015 Test set, for different pixel types in the wide image.}
\small
\renewcommand{\arraystretch}{1.0}
\centering
\begin{tabular}{|c|c|c|c|}
\hline
Model name & Background & Foreground & All Pixels  \\ \hline
SHG-SIDENet &20.19 & 23.44 & 20.73 \\
\hline
MT-TW-SMNet & 15.47& 16.25 & 15.60   \\ \hline
MT-TW Fusion &  11.92 & 12.16 & 11.96  \\ \hline
\end{tabular}
\label{table:2nKITTI}
\end{table}

\section{Tele-Wide Fusion \label{sec:twfusion}}
In Section~\ref{sec:TWDE}, we have introduced multiple ways to solve the tele-wide disparity estimation problem, including TW-SMNet, SHG-SIDENet, and MT-TW-SMNet. In this section, we introduce two  network fusion methods, which are input feature fusion and output feature fusion.
Moreover, we investigate architectures with both input and output fusion. The SIDE network is modified to take the estimated disparity from MT-TW-SMNet as an input feature in addition to the RGB wide input image (RGBD input) which guides the estimation of the inverse depth, and the output of this RGBD SHG-SIDENet is further fused with the output MT-TW-SMNet for more accurate disparity estimation.

\begin{table*}[t!]
\caption{Tele-wide disparity estimation error rates of the input feature fusion network RGBD-SHG-SIDENet using different input sparse disparity maps at the central (cen) and surrounding (sur) regions of the WFOV, on KITTI Stereo 2015 validation dataset. }
\small
\renewcommand{\arraystretch}{1.0}
\centering
\begin{tabular}{|l|c|c|c|c|c|}
\hline
Model name & input-central  & input-sur  & error-all(\%) & error-cen(\%) & error-sur (\%)  \\ \hline
SHG-SIDENet &  N/A & N/A &12.62 & 7.31 & 15.80 \\
TW-SMNet(T) &  N/A & N/A &N/A & \textbf{1.68} & N/A\\
TW-SMNet(W) &  N/A & N/A &13.10 & 1.86 & 19.63 \\
MT-TW-SMNet & N/A & N/A &12.70 & 1.94 & 18.99  \\ \hline
RGBD-SHG-SIDENet(TW(T)) & TW-SMNet(T) & N/A &  11.05 & 5.85 & \textbf{14.15}\\
RGBD-SHG-SIDENet(TW(W)) & TW-SMNet(W) & N/A & \textbf{10.33}  & 2.33 & 15.16 \\
RGBD-SHG-SIDENet(MT) & MT-TW-SMNet & N/A &  10.37 & 3.07  & 14.73 \\
RGBD-SHG-SIDENet(SIDE) & SHG-SIDENet & SHG-SIDENet &  12.42 & 7.23 & 15.29 \\
RGBD-SHG-SIDENet(TW(T), SIDE) & TW-SMNet(T) & SHG-SIDENet & 11.77 & 6.15 & 14.75  \\
RGBD-SHG-SIDENet(TW(W), SIDE) & TW-SMNet(W) & SHG-SIDENet &  10.99  & 3.44 & 15.66   \\
RGBD-SHG-SIDENet(MT, SIDE) & MT-TW-SMNet & SHG-SIDENet & 10.84 & 3.77  & 15.13   \\ \hline
\end{tabular}
\label{table:2NRGBDKITTI}
\end{table*}

\subsection{Input feature fusion: Single image inverse depth estimation with RGBD input}
\begin{figure}
\centering
\includegraphics[width=\linewidth]{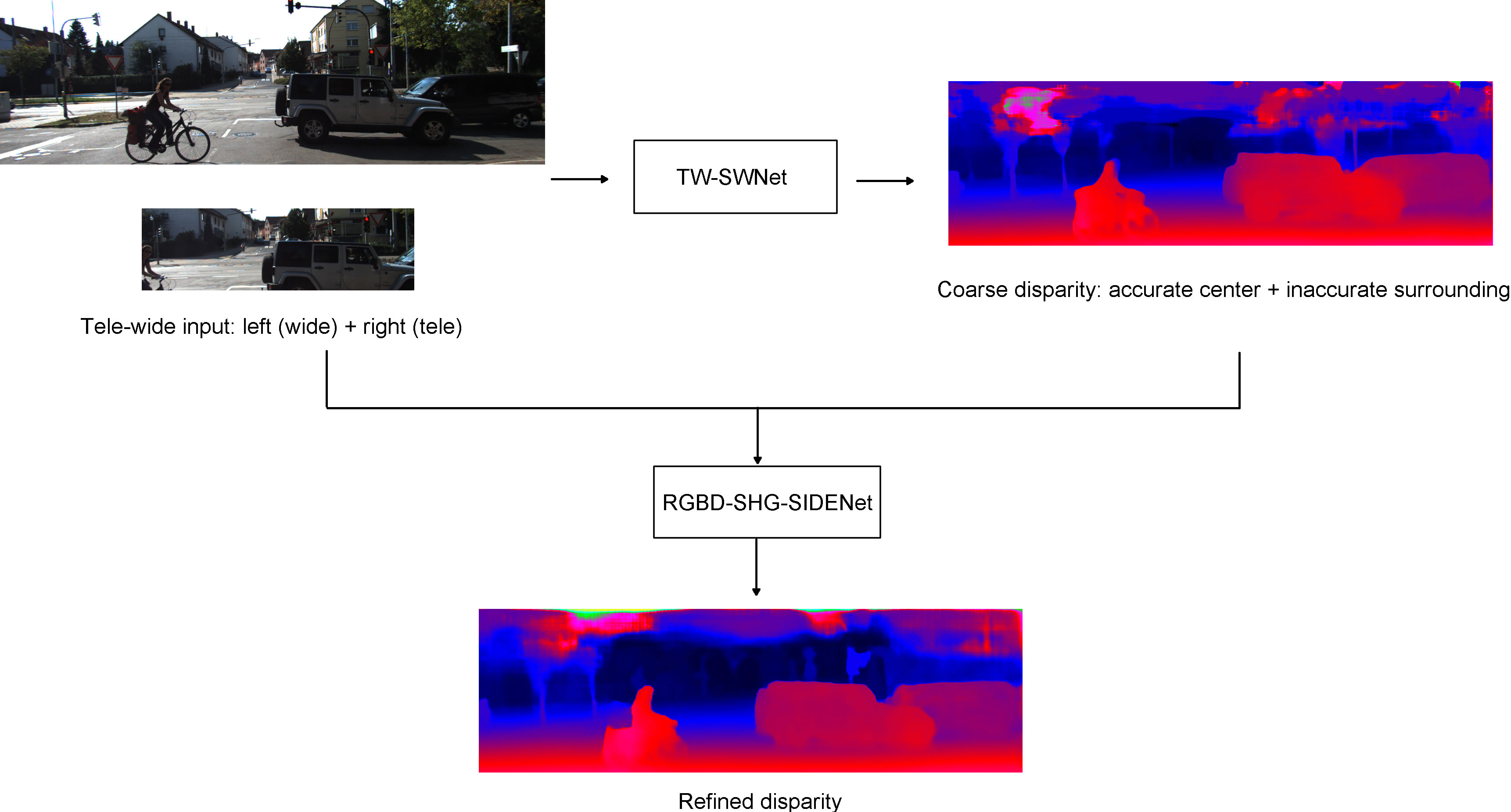}
\caption{Input feature fusion of SHG-SIDENet and TW-SMNet network. }
\label{fig:coarsetofine}
\end{figure}

As a method of input feature fusion, we propose to use the estimated disparity from the stereo matching networks described above  as an input disparity channel to SHG-SIDENet, in addition to the input RGB wide image, to guide the inverse depth estimation of SHG-SIDENet. Since we will adopt the same stacked hourglass architecture for this network as the SHG-SIDENet, we will call this network RGBD-SHG-SIDENet. However, the first layer of the RGBD-SHG-SIDENet is modified from that of the SHG-SIDENet to take the additional input disparity channel. Previous works \cite{ma2017sparse} have shown that using  the knowledge of sparse depth can significantly improve the accuracy of depth estimation, where  it is assumed that such sparse depths can be obtained by some computer vision systems such as simultaneous localization and mapping (SLAM). 
Instead, here we utilize the disparity estimates obtained by tele-wide stereo matching. As shown in Fig.~\ref{fig:coarsetofine}, we first use the TW stereo disparity estimation network to obtain reasonably accurate disparity estimates for the pixels in the tele region. This tele disparity map is concatenated  with the RGB wide image to make the RGBD feature that is input to the RGBD-SHG-SIDENet. Whereas the single image inverse depth estimation network can easily learn the relative distance between objects by scene understanding, the input absolute values can help estimate more accurate inverse depth values. 
The additional disparity input can be generated by  any of the tele-wide disparity estimation networks introduced before. 
RGBD-SHG-SIDE can also be used for input fusion of disparity results from the different SM or SIDE networks, where the disparity map can be constructed from the SM network in the Tele region and from the SIDE network in the surrounding network (by decision selection as in Sec.~\ref{sec:outputfusion}), or multiple disparity maps obtained by different algorithms can be concatenated with the RGB wide image at the RGBD SIDENet. For complexity reduction, the input disparity map can also be estimated using classical stereo matching in the tele region.

\subsection{Experimental results of input feature fusion with RGBD-SHG-SIDENet}
Next, we investigate the effectiveness of input feature fusion using the RGBD-SHG-SIDENet. 
We evaluate the performance on the KITTI Stereo 2015 validation dataset. 
 We train RGBD-SHG-SIDENet by using the sparse ground-truth disparity map  as the additional input disparity channel. 
We conduct an ablation study and compare different options for the input disparity map. At inference time, the additional input disparity channel is sampled from the previously estimated disparity maps by using the same sparsity rate of the ground truth maps used for training. The RGBD-SHG-SIDENet networks could have been alternatively trained directly using the dense or sparse disparity outputs from the networks in Table \ref{table:1nKITTI} as their additional input feature, but this has several drawbacks:   the training convergence is worse, and this requires to train a different model for each possible input which also results in model overfitting and the inability to generalize to different input maps.

Table \ref{table:2NRGBDKITTI} compares the accuracy of the disparity estimated by the RGBD-SHG-SIDENet for different options of construction of the input sparse disparity map. We investigate the different cases where the sparse disparity is sampled from one estimated disparity map. We also investigate the cases where the center disparity and the surrounding disparity are sampled from the outputs of two different networks using decision selection (cf. Sec.~\ref{sec:outputfusion}). The sparse input map is constructed by randomly sampling the estimated disparity maps at rates of $20\%$ of the pixels at the central tele region, and $12\%$ of the pixels at the surrounding region, if it exists.

From Table \ref{table:2NRGBDKITTI}, it can be confirmed that input feature fusion clearly improves the accuracy. Although the accuracy in the central tele region can degrade slightly, the surrounding accuracy improves significantly which improves the overall accuracy. For example, comparing `RGBD-SHG-SIDENet + TW-SMNet(W)' to  `TW-SMNet(W)', the overall error rate is reduced from $13.10\%$ to $10.33\%$, which is even better than the $12.70\%$ attainable from the MT-TW-SMNet. 
 We notice that sampling the central disparity from the stereo matching networks TW-SMNet(T), TW-SMNet(W), or MT-TW-SMNet results in better overall accuracy. Sampling the input disparity map totally from that of the SHG-SIDENet shows the performance attainble using the left wide input image only, and can only slightly improve over the error rate of SHG-SIDENet.
However, it is noticed that the additional sampling of the disparity from the surrounding region estimated by SHG-SIDENet does not improve over only sampling from the central tele regions of the SM disparity maps.  The reason is that the estimated surrounding disparity of SHG-SIDENet is not as accurate as that of the central disparity estimated by stereo matching. Since our training uses the ground truth disparity as the additional input disparity channel, the network expects to get an accurate sparse disparity map at its input.

\begin{table*}[t!]
\caption{Error rate of output feature fusion based on decision selection (DS) and fast global smoothing (FGS) smoothing on the KITTI Stereo 2015 validation dataset. The disparity input of RGBD-SHG-SIDENet comes from TW-SMNet(T).}
\small
\renewcommand{\arraystretch}{1.0}
\centering
\begin{tabular}{|c|c|c|c|c|}
\hline
Model name & type & error-all(\%) & error-cen(\%) & error-sur (\%)  \\ \hline
TW-SMNet(T) &  stereo  &N/A & 1.68 & N/A\\
RGBD-SHG-SIDENet(TW(T)) &  input fusion &11.05 & 5.85 & 14.15 \\ \hline
RGBD-SHG-SIDENet(TW(T)) + TW-SMNet(T) + DS & output fusion & 9.50 & 1.68 & 14.15 \\ 
RGBD-SHG-SIDENet(TW(T)) + TW-SMNet(T) + DS + FGS & output fusion & 9.54 & 1.72 & 14.18 \\ \hline
\end{tabular}
\label{table:2NRFKITTI}
\end{table*}

\subsection{Output feature fusion  \label{sec:outputfusion}}
\begin{figure*}
\centering
\includegraphics[width=0.75\linewidth]{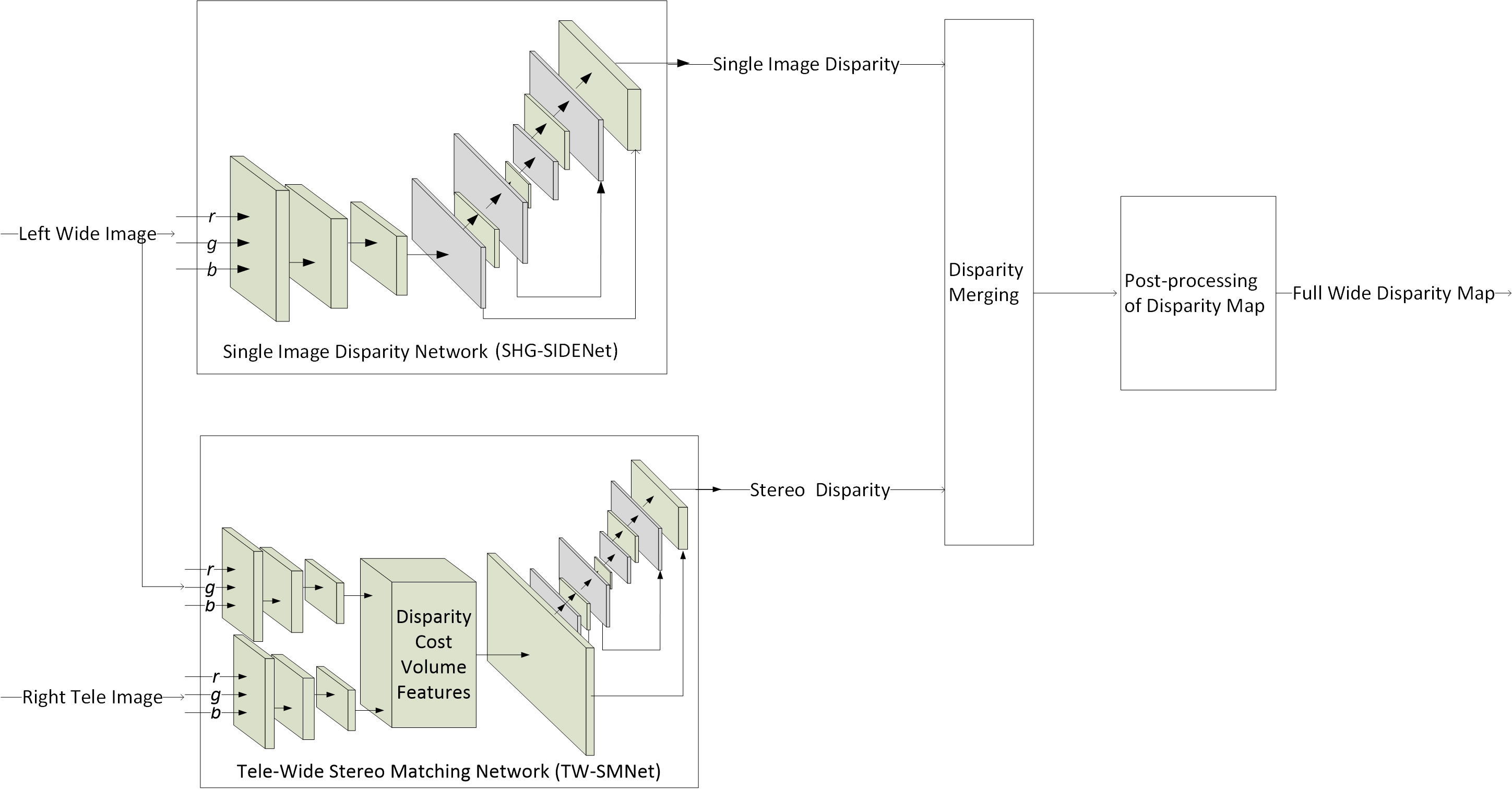}
\caption{Output feature fusion of SHG-SIDENet and TW-SMNet.}
\label{fig:RFSHG-SIDE-RGBTW-SMNet}
\end{figure*}

Fusion of the results from different classifiers has always been an important topic in machine learning \cite{kuncheva2001decision, PAMI_fusion,du2017fused}.
Motivated by this, we investigate methods for the fusion of output features or disparity maps from two or more networks.
The general framework for the proposed output feature fusion is shown in Fig.~\ref{fig:RFSHG-SIDE-RGBTW-SMNet}. To obtain a more accurate disparity map, the output disparity maps from different networks can be merged together by using simple decision selection or by using deep fusion networks or a combination of them. The merged disparity map can be further post-processed to improve its quality,  and reduce visual artifacts for specific applications such as synthesizing the Bokeh effect.

\begin{figure}[h!]
\begin{center}
   \includegraphics[width=\linewidth]{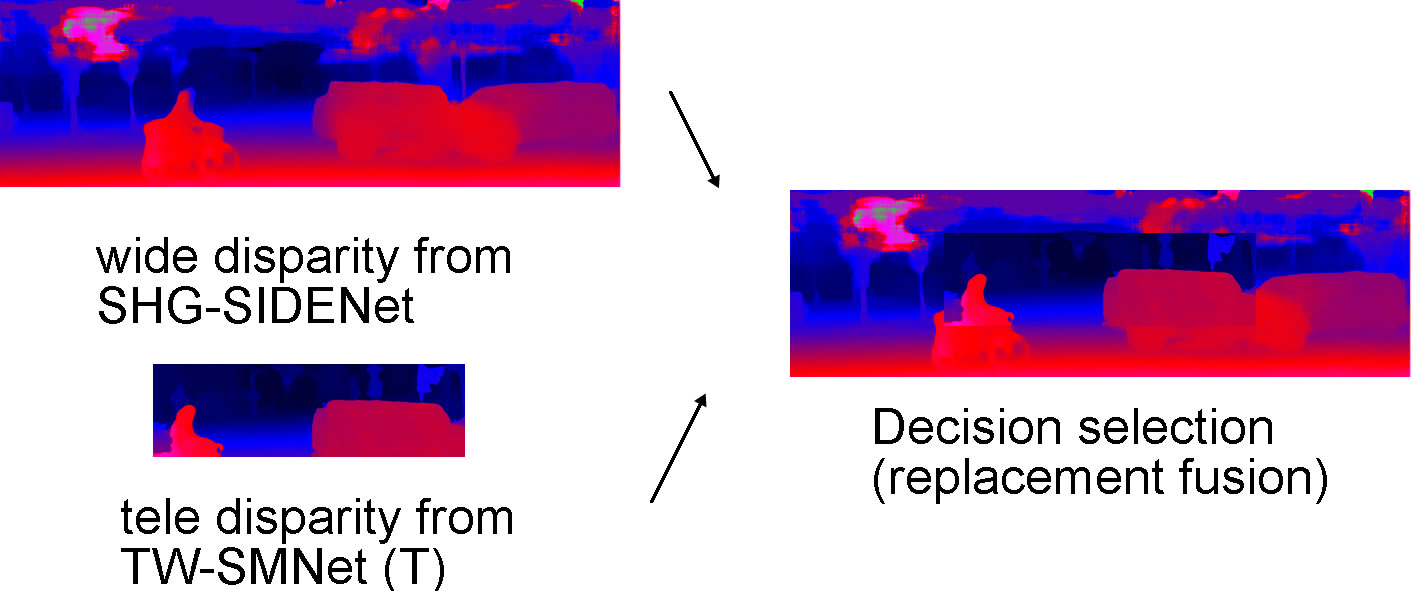}
\end{center}
   \caption{Output feature fusion based on decision selection. The border discontinuity can be observed.}
\label{fig:fgsnopp}
\end{figure}

\subsubsection{Decision selection:} A straightforward solution to obtain a more accurate disparity map from multiple disparity maps is by decision selection. Decision selection attempts to generate a disparity map by selecting the more reliable disparity result at each pixel. We know that the best disparity estimation for the Tele FOV comes from the networks having both the left view and right view as inputs (e.g., TW-SMNet or MT-TW-SMNet). In contrast, since SIDE relies on scene understanding, the best estimation of the surrounding region comes from the SHG-SIDENet. The results of 
Table~\ref{table:1nKITTI} and Table~\ref{table:2NRGBDKITTI}
confirm that the SIDE networks are more accurate in the surrounding regions and that the SM networks give better disparity estimate in the overlapping tele region.  
Hence, to get the best quantitative accuracy, we propose selecting the disparity estimates from one of the TW-SMNets in the central Tele FOV, and selecting it from the SHG-SIDENet in the surrounding region. This will improve the overall disparity estimation accuracy with little additional computational cost.

\begin{figure}[t!]
\centering
\includegraphics[width=\linewidth]{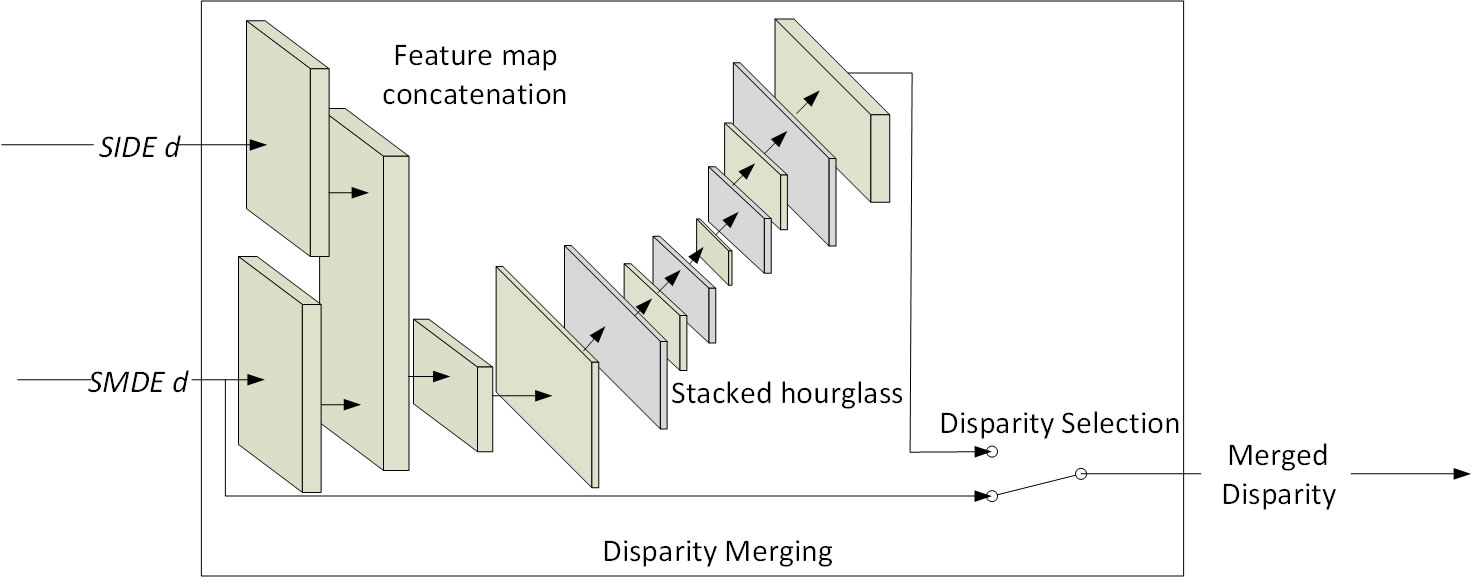}
\caption{Deep network fusion of disparity maps. \label{fig:fusionnet}}
\end{figure}

 Table \ref{table:1nKITTI} and \ref{table:2NRGBDKITTI} show that the best center accuracy is achieved by the TW-SMNet(T), and that the best surrounding accuracy is achieved by RGBD-SHG-SIDENet. So we apply decision selection betwen these two networks, and test the performance on the KITTI Stereo 2015 validation set. The results are shown in Table \ref{table:2NRFKITTI}. As expected, the overall error (`error-all') is reduced from $11.05\%$ to $9.50\%$, due to using the most accurate disparity estimates in both the central and the surrounding regions.

\subsubsection{Post-processing of disparity maps:} Disparity selection is demonstrated in Fig.~\ref{fig:fgsnopp}, which shows that such operation may introduce a large disparity discontinuity along the tele FOV border. Such a disparity discontinuity may be problematic for applications such as synthesizing the Bokeh effect on an image, where the degree of the applied blur depends on the estimated depth. To solve this problem, we utilize a post-processing module to deliver perceptually pleasing disparity maps. One potential way to reduce the effect of the abrupt change in disparity around the tele FOV border is to smooth the disparity map. One solution is edge preserving smoothing using bilateral filters which are based on local averaging~\cite{tomasi1998bilateral}. Another solution is to use the RGB images as a guidance to smoothing, so as to preserve the edges in the RGB image, which is called edge guided filtering. Hence, for post processing we propose using the fast global smoother (FGS) \cite{min2014fast}, which optimizes a global objective function defined with data constraints and a smoothness prior. Since the FGS filtered values around the tele FOV border depends on the whole disparity map, we calculate the filtered values around the border using the global filters by deploying FGS. To smooth the transition in disparity values between the central and surrounding region, only the strip around the boundary in the merged disparity map is replaced with the filtered one.

\begin{figure*}[t!]
\centering
\includegraphics[width=0.7\linewidth]{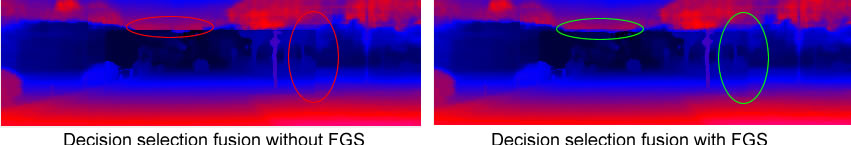}
\caption{Feature output fusion using decision selection fusion (left) 
and with additional post-processing using FGS smoothing (right).}
\label{fig:fgsbs}
\end{figure*}

As shown in the last row of Table \ref{table:2NRFKITTI}, the error rate increases 0.04\% by applying FGS due to the smoothing around the tele FOV border. However,  the perceptual quality improved significantly as the disparity discontinuity around the border is smoothed out, as shown the Fig.~\ref{fig:fgsbs}.

\begin{figure*}
\centering
\includegraphics[width=\linewidth]{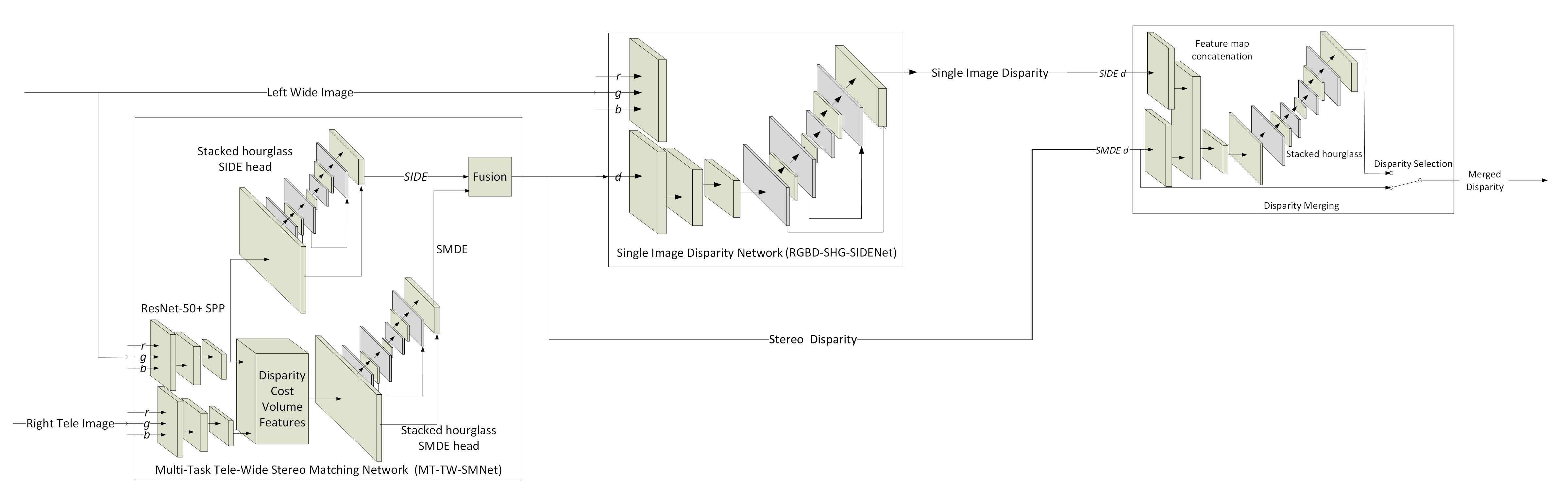}
\caption{ MT-TW Fusion: Tele-wide disparity estimation using both input feature fusion and output feature fusion, which has the best performance among the tested architectures. The disparity merging module performs deep network fusion using a stacked hourglass network, followed by decision selection. }
\label{fig:RGBDfusion}
\end{figure*}

\subsubsection{Deep network fusion of disparity maps:} Above we observed that the fusion of multiple disparity maps with simple methods  such as decision selection, often requires post processing with global smoothers to reduce the fusion artifacts. 
That motivated us to design deep neural networks to fuse the disparity maps generated by different schemes. 
Deep network fusion of multiple output features has been investigated before in different contexts, such as  the fusion of the output images from different super-resolution networks \cite{ren2017image}. Deep network fusion is characterized by its capability of taking into account both the global and local contexts, as defined by the deep network's receptive field. 

 The proposed disparity fusion network concatenates two estimated disparity maps at its input, and outputs a refined disparity map, as shown in  Fig.~\ref{fig:fusionnet}. In our experiments, one of the input maps is obtained by SIDE and the other input disparity map is obtained by one of the stereo-matched disparity estimation (SMDE) techniques. The fusion network is trained using the classification-based robust regression loss given by \eqref{eq:loss}, as used for training our TW-SMNets. For simplicity, we utilized the stacked hourglass head used in the TW-SMNet and SHG-SIDE as the fusion network. Hence, the proposed fusion is called SHG-Fusion.Compared to decision selection, deep network fusion of the disparity maps achieves better accuracy. Moreover, the deeply fused disparity maps do not need FGS post-processing as they do not suffer from disparity discontinuity around the tele FOV border.

\begin{figure*}[t!]
\centering
\includegraphics[width=\linewidth]{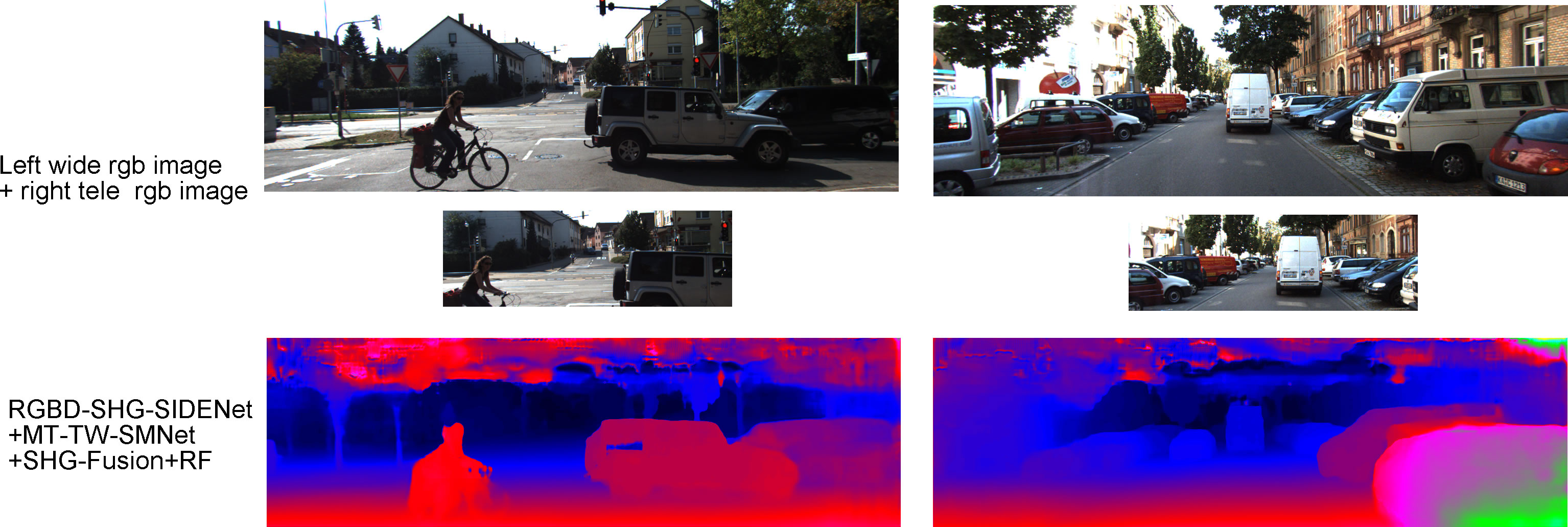}
\caption{Example outputs of our estimated tele-wide disparity from the TW-SM Fusion network. }
\label{fig:best}
\end{figure*}

 \begin{table*}
\caption{Error rate of output feature fusion by disparity fusion network between RGBD-SHG-SIDENet and MT-TW-SMNet in KITTI validation set. `cen' stands for `center', and `sur' stands for `surrounding'. }
\small
\renewcommand{\arraystretch}{1.0}
\centering
\begin{tabular}{|c|c|c|c|c|}
\hline
Model name  & error-all(\%) & error-cen(\%) & error-sur (\%)  \\ \hline
MT-TW-SMNet & 12.70 & 1.94 & 18.99  \\
RGBD-SHG-SIDENet(MT) &  10.37 & 3.07  & 14.73 \\ \hline
RGBD-SHG-SIDENet(MT) + DS & 10.15 & 1.94 & 14.73  \\ 
RGBD-SHG-SIDENet(MT) + MT-TW-SMNet + SHG & 9.84 & 2.50 & 14.24 \\ 
RGBD-SHG-SIDENet(MT) + MT-TW-SMNet + SHG + DS & 9.64 & 1.94 & 14.24  \\ \hline
\end{tabular}
\label{table:3Nfusion}
\end{table*}

\begin{table*}
\caption{End-point error of output feature fusion by disparity fusion network between RGBD-SHG-SIDENet and MT-TW-SMNet in SceneFlow test set. `cen' stands for `center', and `sur' stands for `surrounding'.}
\small
\renewcommand{\arraystretch}{1.0}
\centering
\begin{tabular}{|c|c|c|c|c|}
\hline
Model name  & error-all(pixels) & error-cen(pixels) & error-sur (pixels)  \\ \hline
RGBD-SHG-SIDENet(MT) &6.39 & 8.22 & 6.13 \\
MT-TW-SMNet&5.61 & 1.62 & 7.08  \\ \hline
RGBD-SHG-SIDENet(MT) + MT-TW-SMNet + SHG &  5.50 & 3.58 & 6.10 \\ 
RGBD-SHG-SIDENet(MT) + MT-TW-SMNet + SHG + DS &  5.33 & 1.62 & 6.10  \\ \hline
\end{tabular}
\label{table:3NfusionSF}
\end{table*}

\subsubsection{Combining the different fusion methods:} 
As shown in Fig.~\ref{fig:fusionnet}, decision selection can be further applied between the deeply fused disparity map at the output of the fusion network and the  disparity map input to the fusion network which was obtained by stereo matching. The crux of such additional merging is that the fusion network helps improve the overall disparity by especially improving the estimated disparity in the surrounding region, but the stereo matched disparity maps at the input may still be more accurate in the tele FOV region.

Next, we give some experimental results with output feature fusion. In Table \ref{table:3Nfusion}, we find that the error rate can be reduced from $10.37\%$ to $9.85\%$ by the stacked hourglass fusion network, and to $9.64\%$ by further applying decision selection with MT-TW-SMNet.  Although the accuracy of stacked hourglass fusion is slightly lower than RGBD-SIDENet based on TW-SMNet (T) in Table \ref{table:2NRFKITTI}, its output does not have boundary discontinuity due to the using of fusion network.
The accuracy of decision-level fusion of SHG-SIDENet and MT-TW-SMNet on the Sceneflow dataset is shown in Table \ref{table:3NfusionSF}. From Table \ref{table:3NfusionSF}, we observe that by using SHG-fusion, the end-point-error is reduced from $5.61$ pixels to $5.50$ pixels, and further to $5.31$ pixels by decision selection. These results confirm that our proposed fusion is effective in improving the accuracy of the tele-wide disparity estimation network.

Both output feature fusion and input feature fusion can also be combined together in the same scheme, as demonstrated in Fig.~\ref{fig:RGBDfusion}. For example, the SIDE map can be obtained by the RGBD-SHG-SIDENet which samples the stereo-matching output of the MT-TW-SMNet disparity map to construct a sparse disparity feature, that is fused at the input of the RGBD SIDE network. The disparity maps from the same stereo-matching branch of the MT-TW-SMNet and from the output of the RGBD-SHG-SIDENet are then deeply fused using SHG-Fusion. This architecture gave the best accuracy among the tele-wide disparity estimation methods, we tested.

We submitted our best results to the KITTI leaderboard server. Our best performance on the KTTI test set on the leaderboard is by the fusion: RGBD-SHG-SIDENet (D comes from the estimated tele disparity from MT-TW-SMNet) + MT-TW-SMNet + SHG-fusion + decision selection. KITTI leaderboard utilizes a different evaluation protocol as the validation set, where the error of foreground pixel (error-fg) and the error of background pixel (error-bg) are calculated. This performance is shown in Table~\ref{table:2nKITTI} as `MT-TW Fusion'. Although this accuracy is not as good as the top ranked  submissions, it only utilizes the  left wide image and  the right tele image, and uses much less information than other submissions which use the full wide left and right images. Our proposed tele-wide stereo matching fusion scheme still ranks better than recent schemes for wide-wide stereo matching that also use the interframe optical flow information \cite{schuster2018combining}. Example disparity outputs from the MT-TW Fusion network on test examples are given in Fig.~\ref{fig:best}, and more detailed visualizations can be found at \cite{TWSMNet-kitti}.

\subsection{Post-processing for Bokeh synthesis}

\begin{figure}[t!]
\centering
\includegraphics[width=\linewidth]{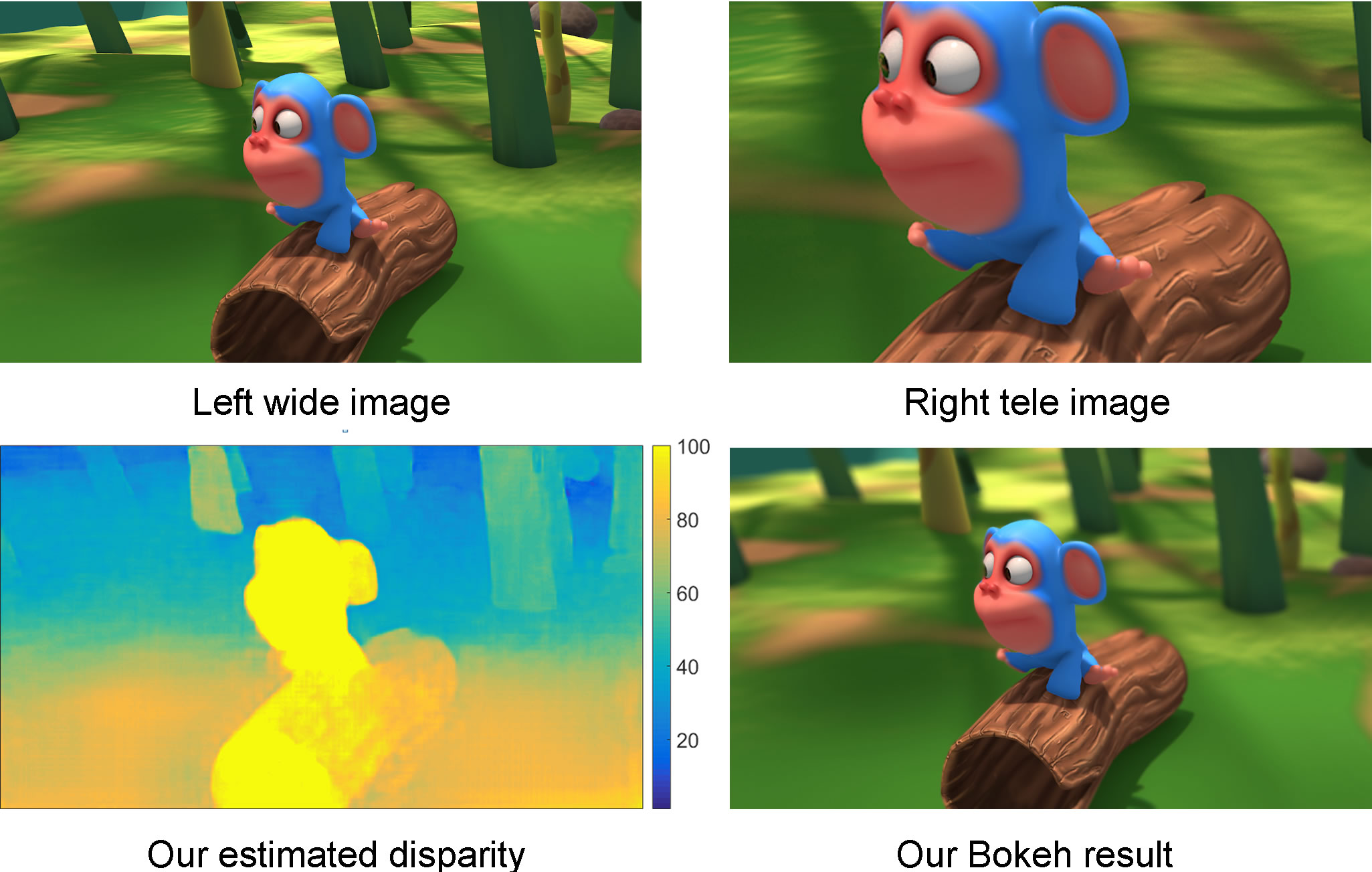}
\caption{Examples of the image Bokeh results using tele-wide disparity on SceneFlow dataset. }
\label{fig:BokehSFRes}
\end{figure}

Given a focused image of a scene and the depth map of the scene, the Bokeh effect can be synthesized by post-processing to render blurry out-of-focus areas in the image \cite{mcgraw2015fast}. 
We use the estimated tele-wide depth map to synthesize the Bokeh effect on the full WFOV. One problem is that there are some pixels which are not correctly estimated, especially at the top (sky) regions of KITTI images due to the lack of ground-truth labels. So we propose an additional post-processing module to remove these incorrect pixels before using it for Bokeh. We know that image Bokeh requires an input focus point to locate the desired foreground object. A feasible way to remove the bad estimated pixels is to suppress the regions with high disparity that don't include the focus point. So we design the following algorithm based on the connected components and distance transform, as shown in Fig.~\ref{fig:ppBokeh}. After  locating the foreground region by the Bokeh focus point, we suppress all other disparities relative to their distance from the foreground region.  Examples for the Bokeh effect synthesized from input tele-wide stereo image pairs are shown for the KITTI Stereo 2015 and the SceneFlow datasets in Fig.~\ref{fig:tele_wide_kitti2} and Fig.~\ref{fig:BokehSFRes}, respectively.

 \begin{algorithm}
  \caption{Post-processing the estimated disparity for image Bokeh. \label{fig:ppBokeh}}
  \begin{algorithmic}[1]
  \renewcommand{\algorithmicrequire}{\textbf{Input:}}
  \renewcommand{\algorithmicensure}{\textbf{Output:}}
  \REQUIRE Wide disparity $d$, Focus point $(i,j)$
  \ENSURE  Post-processed disparity map  $d'$
   \STATE Define a pixel $(x,y)$ as foreground if and only if $0.7 d(i,j) \leq d(x,y) \leq 1.3 d(i,j)$	
	 \STATE Extract a disparity map $d_f=d$ for all foreground pixels, and $d_f=0$ otherwise.
	 \STATE Generate the foreground mask $b_f=\{d_f \neq 0\}$.
	 \STATE Find $n$ connected components $\{C_1,C_2,…,C_n \}$ in the $b_f$.
	 \STATE Find the connected component $C_t$ that includes the focus point $(i,j)$.
   \FOR {$k = 1$ to $n$, and $k \ne t$} 
   \FOR {pixels $(x,y) \in C_k$, and $i \ne 0$ }
   \STATE Calculate the closet Euclidean distance $p(x,y)$ to any pixel not in $\{C_1,C_2,…,C_n \}$
	 \STATE $d'(x,y)=d(x,y)\frac{1 - 2 p(x,y)}{p(x,y)}$ 
   \ENDFOR
  \ENDFOR
  \RETURN $d'$ 
  \end{algorithmic} 
  \end{algorithm}

\section{Conclusion \label{sec:conc}}
In this paper, we introduced the tele-wide stereo matching problem. We established baseline solutions to estimate the inverse depth for the full wide field of view (FOV) from a tele-wide stereo image pair having a left wide FOV (WFOV) image and a right tele FOV (TFOV) image. To improve the estimation accuracy, we further introduced a multitask tele-wide stereo matching  network (MT-TW-SMNet) that is trained end-to-end to learn both a disparity estimation objective and an inverse depth estimation objective. 
We showed that input feature fusion with RGBD inverse depth estimation can significantly improve the depth estimation quality in the regions surrounding the TFOV, where the stereo information is missing.
We further explored different output feature fusion methods, such as with decision selection between the proposed stereo matching and inverse depth estimation techniques, or with context-aware deep network fusion using stacked hour glass networks. 
Experimental results on KITTI and SceneFlow datasets demonstrate that our proposed approaches achieve considerable performance in the tele-wide disparity estimation scheme, and perform better than other popular methods that perform stereo matching between two wide FOVs. We demonstrate the usefulness of this approach, by synthesizing the Bokeh effect on the full WFOV, when the input is a tele-wide stereo image pair.  Although not considered in this paper, our work can also be generalized to systems with more than two cameras or multi-camera systems with different focal lengths, by estimating the union FOV for one stereo pair at a time, and recursively merging the results from the different stereo pairs to estimate the depth map for the union FOV of all cameras.

{\small
\bibliographystyle{IEEEtran}
\bibliography{TWbib}
}

\end{document}